\documentclass[lettersize,10pt,journal,twoside]{IEEEtran}
\usepackage{ieee_robotics}
\acrodef{dof}[DoF]{Degrees of Freedom}
\acrodef{ramis}[RAMIS]{robot-assisted minimally invasive surgery}
\acrodef{fov}[FOV]{field of view}
\acrodef{mlp}[MLP]{multilayer perceptron}
\acrodef{mse}[MSE]{mean squared error}
\acrodef{svm}[SVM]{support vector machine}

\newcommand{\sensorname}{MiniTac\xspace}

\usepackage{orcidlink}
\usepackage[detect-weight,mode=text]{siunitx}
\title{\LARGE \bf \sensorname: An Ultra-Compact \SI{8}{mm} Vision-Based Tactile Sensor for\\Enhanced Palpation in Robot-Assisted Minimally Invasive Surgery}

\author{Wanlin Li\orcidlink{0000-0002-8538-8571}, Zihang Zhao\orcidlink{0000-0003-3215-7152}, Leiyao Cui\orcidlink{0009-0009-4925-6983}, Weiyi Zhang\orcidlink{0000-0003-4178-7084}, Hangxin Liu\orcidlink{0000-0002-3003-8611}, Li-An Li\orcidlink{0000-0001-8628-4101}, and Yixin Zhu\orcidlink{0000-0001-7024-1545}

\thanks{Manuscript received: June 12, 2024; Revised: September 13, 2024. Accepted October 17, 2024.}
\thanks{This paper was recommended for publication by Editor Ashis Banerjee upon evaluation of the Associate Editor and Reviewers’ comments.}
\thanks{This work is supported in part by the National Science and Technology Major Project (2022ZD0114900), the National Natural Science Foundation of China (62376009), the Beijing Nova Program, the State Key Lab of General AI at Peking University, the PKU-BingJi Joint Laboratory for Artificial Intelligence, and the National Comprehensive Experimental Base for Governance of Intelligent Society, Wuhan East Lake High-Tech Development Zone. (W. Li, Z. Zhao, L. Cui, and W. Zhang contributed equally to this work. Corresponding authors: Yixin Zhu, and Li-An Li.)
}%
\thanks{Wanlin Li and Hangxin Liu are with the State Key Laboratory of General Artificial Intelligence, Beijing Institute for General Artificial Intelligence (BIGAI), Beijing 100080, China.}
\thanks{Zihang Zhao and Yixin Zhu (e-mail: yixin.zhu@pku.edu.cn) are with the Institute for Artificial Intelligence, Peking University, Beijing 100871, China.}
\thanks{Leiyao Cui is with Shenyang Institute of Automation, Chinese Academy of Sciences, Shenyang 110169, China, and interned at the Institute for Artificial Intelligence, Peking University, Beijing 100871, China.}
\thanks{Weiyi Zhang and Li-An Li (e-mail: lila@ieee.org) are with the PLA General Hospital, Beijing 100853, China.}
\thanks{Digital Object Identifier (DOI): 10.1109/LRA.2024.3487516}}

\begin{document}

\maketitle

\begin{abstract}
\Ac{ramis} provides substantial benefits over traditional open and laparoscopic methods.
However, a significant limitation of \ac{ramis} is the surgeon's inability to palpate tissues, a crucial technique for examining tissue properties and detecting abnormalities, restricting the widespread adoption of \ac{ramis}.
To overcome this obstacle, we introduce \sensorname, a novel vision-based tactile sensor with an ultra-compact cross-sectional diameter of 8 mm, designed for seamless integration into mainstream \ac{ramis} devices, particularly the Da Vinci surgical systems.
\sensorname features a novel mechanoresponsive photonic elastomer membrane that changes color distribution under varying contact pressures. This color change is captured by an embedded miniature camera, allowing \sensorname to detect tumors both on the tissue surface and in deeper layers typically obscured from endoscopic view. \sensorname's efficacy has been rigorously tested on both phantoms and ex-vivo tissues.
By leveraging advanced mechanoresponsive photonic materials, \sensorname represents a significant advancement in integrating tactile sensing into \ac{ramis}, potentially expanding its applicability to a wider array of clinical scenarios that currently rely on traditional surgical approaches.
\end{abstract}

\begin{IEEEkeywords}
Robot-assisted minimally invasive surgery, miniature tactile sensor, tumor detection
\end{IEEEkeywords}

\section{Introduction}

\IEEEPARstart{R}{obot-assisted} minimally invasive surgery (RAMIS) represents a pivotal advancement in surgical methodologies, providing a plethora of benefits. It not only ensures smaller incisions, accelerated patient recovery, diminished postoperative pain, and a reduced risk of infection compared to traditional open surgeries but also surpasses laparoscopic surgeries in terms of enhanced 3D visualization and superior precision and control over surgical instruments~\cite{kwok2022soft,vilsan2023open}.

Despite these advantages, a critical limitation of \ac{ramis} is the absence of tactile feedback, which restricts the surgeon's ability to perform palpation~\cite{liu2011rolling,kwok2022soft,vilsan2023open}. In both open and laparoscopic surgeries, surgeons palpate the patient's soft tissues using their fingers or surgical probes to identify underlying anatomical structures invisible to the naked eye or endoscopic cameras~\cite{liu2011rolling}. The tactile feedback from this kind of manipulation is crucial for accurately localizing tumors, as solid tumors can be distinguished from normal tissue by their increased firmness~\cite{vaccarella2016worldwide,riggi2021ewing}, which is vital for ensuring minimal resection margins~\cite{kwok2022soft}. The lack of this capability in \ac{ramis} significantly curtails its application across a broader range of surgical scenarios~\cite{gerald2024soft}.

\begin{figure}[t!]
    \centering
    \begin{minipage}{.63\linewidth}
        \begin{subfigure}[b]{\linewidth}
            \includegraphics[width=\linewidth]{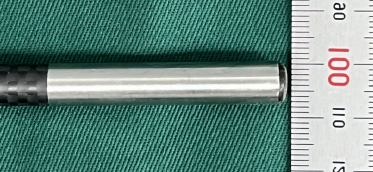}
            \caption{}
        \end{subfigure}\\%
        \begin{subfigure}[b]{\linewidth}
            \includegraphics[width=\linewidth]{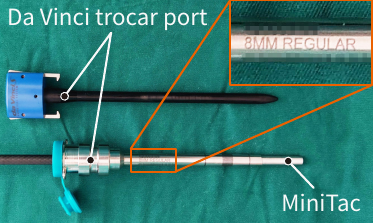}
            \caption{}
        \end{subfigure}
    \end{minipage}%
    \hfill%
    \begin{minipage}{.36\linewidth}
        \begin{subfigure}[b]{\linewidth}
            \includegraphics[width=\linewidth]{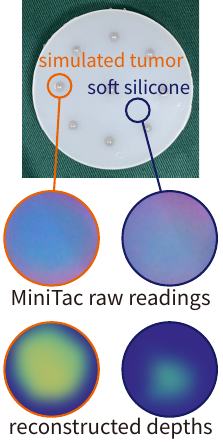}
            \caption{}
        \end{subfigure}
    \end{minipage}
    \caption{\textbf{\sensorname is compatible with Da Vinci robotic surgical systems for palpation.} (a) \sensorname features an ultra-compact design with a cross-sectional diameter of \SI{8}{mm}, suitable for standard trocar ports. (b) Its design ensures seamless integration with Da Vinci systems, commonly used in \ac{ramis}. (c) \sensorname provides high spatial resolution and sensitivity, which facilitates detailed deformation information at the contact site, enabling precise detection of hard tumors within soft, simulated normal tissue.}
    \label{fig:teaser}
\end{figure}

In addressing this limitation, there has been significant interest in integrating tactile sensors into \ac{ramis} systems~\cite{konstantinova2014implementation,othman2022tactile}. However, the trocar port size---typically an 8 mm diameter circle, as is standard in the widely-used Da Vinci robotic surgical systems~\cite{Davinci}---presents a unique challenge. This challenge involves developing a sensor that not only provides high-density tactile sensing but is also compact enough for insertion through the port. Despite extensive research, current tactile sensing solutions continue to struggle with achieving miniaturization while maintaining a sufficient number of taxels (sensing units) within the limited space (reviewed in detail in \cref{sec:relatedwork}). This challenge has hindered the integration of tactile sensors into \ac{ramis} systems and limited their overall impact. The lack of a commercially available solution further highlights the complexity of this challenge.

In this paper, we introduce \sensorname, a novel vision-based tactile sensor specifically designed for \ac{ramis} systems. The ultra-compact sensor has a diameter of \SI{8}{mm} (\cref{fig:teaser}(a)), ideally suited to fit through standard trocar ports of Da Vinci robotic surgical systems (\cref{fig:teaser}(b)). \sensorname features a mechanoresponsive photonic elastomer membrane that alters its color distribution in response to different levels of contact~\cite{miller2022scalable}. This altered color distribution is captured by a miniature camera at a high spatial resolution of \SI{10}{\um}, endowing the device with sensitivity akin to having \num{300000} taxels across its surface. This enables \sensorname to accurately map detailed deformations at the contact site and reliably identify simulated embedded tumors within the contact region, as depicted in \cref{fig:teaser}(c).

The two principal contributions of this work are as follows:
\begin{itemize}[leftmargin=*,noitemsep,nolistsep]    
    \item We present \sensorname, an innovative design for miniature tactile sensors with high-density taxels. This design offers \ac{ramis} systems nuanced tactile sensing capabilities. As a result, it becomes feasible to palpate tissue surface for tumor detection during \ac{ramis} procedures.
    \item We demonstrate the functionality of \sensorname using both phantoms and ex-vivo tissue samples. Our tests show clear distinctions in sensor readings between normal and pathological tissue areas, confirming the \sensorname's effectiveness.
\end{itemize}

The remainder of the paper is organized as follows: \cref{sec:relatedwork} reviews relevant literature. \cref{sec:sensor} introduces \sensorname, detailing its design objectives, fabrication processes, calibration methods, and sensitivity. \cref{sec:exps} validates the capabilities of \sensorname in detecting hard embedded tumors in both phantom models and ex-vivo tissue. In \cref{sec:discussion}, we discuss the unique features and advantages of \sensorname, emphasizing its potential to enhance palpation-equivalent capabilities in \ac{ramis} as compared to other tactile sensors. We conclude with key findings and implications in \cref{sec:conclusion}.

\section{Related Work}\label{sec:relatedwork}

Research into enhancing \ac{ramis} systems with palpation capabilities continues amid substantial general progress within the field of \ac{ramis}. Traditionally, surgeons depend on advanced stereo vision feedback as offered by \ac{ramis} systems such as the Da Vinci robotic surgical system~\cite{kwok2022soft}. These systems enable surgeons to infer tissue characteristics by observing the deformations resulting from contact. However, this method demands extensive training, and the accuracy of the outcomes can vary significantly depending on the surgeon's experience level. As a result, there is a growing desire among surgeons for the integration of augmented haptic feedback.

Tactile sensors designed to measure contact properties have shown promise in various robotic applications. However, their integration into \ac{ramis} systems poses unique challenges, principally due to the confined operational space. The sensing technologies currently used in \ac{ramis} are typically piezoresistive~\cite{kalantari2010design}, piezoelectric~\cite{lee2014micro}, capacitive~\cite{kim2018sensorized}, or optical-fiber-based ~\cite{xie2014optical}. These technologies generally provide measurements at a single or a few points, limiting spatial resolution and effectiveness for tasks requiring the examination of large areas. Although rolling mechanisms have been developed to extend coverage~\cite{liu2011rolling}, they increase the complexity and duration of surgical procedures.

Emerging technologies in electronic skins (E-skins) leverage advancements in materials science, nanotechnology, and electronics to offer flexible, high-resolution, and multi-modal tactile sensing~\cite{luo2023technology}. Despite their potential, E-skins are known to have durability issues and their production is subject to complex manufacturing processes and high costs.

Vision-based tactile sensors utilize an embedded camera to capture optical changes of a front soft membrane, decoding its deformation and providing tactile feedback. These sensors offer extremely high spatial resolution at the pixel level, straightforward manufacturing processes, and low costs. Based on their working principles, they can be broadly categorized into photometric stereo types~\cite{yuan2017gelsight,Gelsight-mini,lambeta2020digit,taylor2022gelslim,li20233}, darkness mapping types~\cite{lin20239dtact}, binocular stereo types~\cite{zhang2023gelstereo}, and compound-eye stereo types~\cite{luo2023compdvision}. Extensive research has demonstrated that these sensors can precisely detect object texture, hardness, and surface roughness, closely mirroring the efficacy of manual palpation~\cite{yuan2016estimating,yan2022surface,zhao2024tac}. Indeed, their effectiveness in identifying simulated embedded tumors has been demonstrated, highlighting their potential in medical applications~\cite{jia2013lump,cui2023prototype,di2024using}. However, these sensors tend to be bulky within the context of \ac{ramis}, posing significant challenges for their integration into \ac{ramis} systems. The characteristics of these sensors are summarized in \cref{tab:compare_with_other_sensors}.

Structurally colored materials present a promising approach for developing miniature vision-based tactile sensors due to their minimal illumination requirements~\cite{yue2015tunable, cho2015elastoplastic, luo2023technology, miller2022scalable}. Our research capitalizes on one type of structurally colored materials called mechanoresponsive photonic materials~\cite{miller2022scalable} and introduces a concise and effective illumination strategy, facilitating the miniaturization of vision-based tactile sensors to millimeter scales while ensuring extensive coverage of the effective measurement area. Our device, termed \sensorname, replaces traditional surround-illumination systems with a simplified white lighting setup positioned directly above the sensor's outermost layer. This innovative design enables the compact construction of vision-based sensors without sacrificing the scope of sensing regions. The performance of \sensorname in detecting tumors has been rigorously confirmed through tests on both phantoms and ex-vivo tissues.

\begin{table}[ht!]
    \small
    \centering
    \setlength{\tabcolsep}{1.5pt}
    \caption{\textbf{Comparison of compact high-resolution tactile sensors: P.S. (Photometric Stereo), D.M. (Darkness Mapping), B.S. (Binocular Stereo), C.S. (Compound-Eye Stereo), and M.P.M. (Mechanoresponsive Photonic Material).}}
    \label{tab:compare_with_other_sensors}
    \resizebox{\linewidth}{!}{%
        \rowcolors{2}{Ivory3}{white}
        \begin{tabular}{lccccc}
            \toprule
            Sensor & \makecell[c]{Working \\ Principle} & {\makecell[c]{Illumination \\ (Channel)}} & {\makecell[c]{Dimension \\ ($\mathrm{mm}^2$)}$\downarrow$} & {\makecell[c]{Sensing Region \\ ($\mathrm{mm}^2$)}}  & SR/D$\uparrow$ \\ 
            \hline
            GelSight~\cite{yuan2017gelsight} & P.S. & $3-$RGB & $35\times35$   &  $18\times14$ & $0.21$\\
            GelSight-Mini~\cite{Gelsight-mini} & P.S. & $3-$RGB & $32\times28$ & $19\times15$ & $0.32$\\
            DIGIT~\cite{lambeta2020digit} & P.S. & $3-$RGB & $20\times27$ & $19\times16$ & $0.56$\\
            GelSlim3.0~\cite{taylor2022gelslim} & P.S. & $3-$RGB & $80\times37$ & $675$ & $0.23$\\
            L\(^3\) F-TOUCH~\cite{li20233} & P.S. & $3-$RGB & $40\times25$ & $20\times14$ & $0.28$\\
            9DTact~\cite{lin20239dtact} & D.M. & $1-$White & $32.5\times25.5$ & $24\times18$ & $0.52$\\
            GelStereo2.0~\cite{zhang2023gelstereo} & B.S. & $1-$White & $30\times30$ & $23\times23$ & $0.59$\\
            CompdVision~\cite{luo2023compdvision} & C.S. & $1-$White & $22\times14$ & $170$ & $0.55$\\
            \textbf{\sensorname} &{\makecell[c]{M.P.M.}}  & $1-$White & $\boldsymbol{\pi \times 4^2}$ & $\pi \times 3.5^2$ & $\textbf{0.77}$ \\
            \hline
        \end{tabular}%
    }%
\end{table}

\section{The \sensorname Sensor}\label{sec:sensor}

This section outlines the developmental process for \sensorname, starting with design specifications that would ensure compatibility with \ac{ramis}, as detailed in \cref{sec:design_specs}. We then describe the design and fabrication processes developed to meet these specifications---see \cref{sec:design_details}. This leads to a discussion about the calibration method employed to correlate raw sensor readings with deformation data at the contact site, as outlined in \cref{sec:calibration}. Finally, we test \sensorname's sensitivity, as detailed in \cref{sec:sensor_sensitivity}, and its repeatability and hysteresis in \cref{sec:sensor_hysteresis}.

\subsection{Design Objectives and Specifications}\label{sec:design_specs}

The primary objective is to facilitate effective robotic palpation during \ac{ramis} for tumor detection. A critical requirement is that the sensor has to fit through trocar ports typically used in robotic surgical systems. Its cross-sectional diameter therefore has to be within the dimensions of the trocar's diameter, which, using the Da Vinci system as a guideline, means it can be no more than \SI{8}{mm}.

Apart from size constraints, the ideal sensor must be highly sensitive to variations in deformation, as tumors are generally harder than the normal surrounding tissue~\cite{yuan2016estimating}. It should also provide high spatial resolution to enhance the efficacy of palpation. A further consideration is that any parts of the tactile sensor that come into contact with human tissue must be constructed from soft materials to minimize the risk of injury during palpation. 

\subsection{Design and Fabrication Details}\label{sec:design_details}

\begin{figure}[t!]
    \centering
    \begin{subfigure}[b]{\linewidth}
        \includegraphics[width=\linewidth]{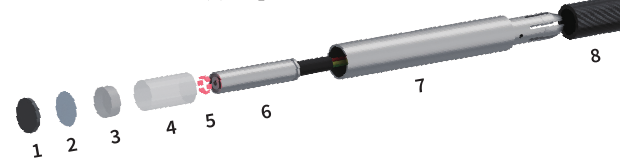}
        \caption{an exploded view}
    \end{subfigure}\\%
    \begin{subfigure}[b]{\linewidth}
        \includegraphics[width=\linewidth]{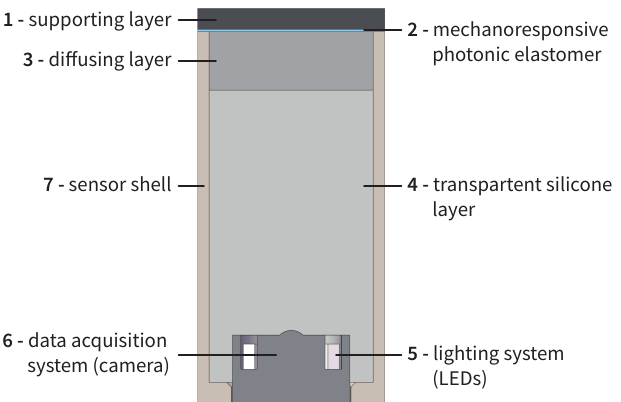}
        \caption{a schematic diagram}
    \end{subfigure}\\%
    \caption{ \textbf{Mechanical design of \sensorname. }(a) An exploded view of \sensorname, highlighting its components. Notably, component 8 is a carbon cube acting as an extended handle to ensure compatibility with \ac{ramis}. (b) A schematic diagram of the fully assembled \sensorname is presented, showing how the components fit together.}
    \label{fig:sensor_design}
\end{figure}

We carefully design and manufacture the \sensorname sensor to meet the given objectives and specifications, as illustrated in the exploded view and schematic diagram---see \cref{fig:sensor_design}(a) and \cref{fig:sensor_design}(b). Consideration is given to the design and selection of \sensorname's components, as each element will have an impact on the overarching goal. 
 
\paragraph*{Mechanoresponsive photonic membrane and its supporting layer}

\begin{figure}[t!]
    \centering
    \begin{subfigure}[b]{.25\linewidth}
        \centering
        \includegraphics[width=.85\linewidth]{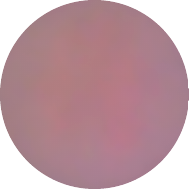}
        \caption{}
    \end{subfigure}%
    \begin{subfigure}[b]{.25\linewidth}
        \centering
        \includegraphics[width=.85\linewidth]{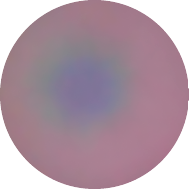}
        \caption{}
    \end{subfigure}%
    \begin{subfigure}[b]{.25\linewidth}
        \centering
        \includegraphics[width=.85\linewidth]{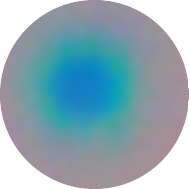}
        \caption{}
    \end{subfigure}%
    \begin{subfigure}[b]{.25\linewidth}
        \centering
        \includegraphics[width=.85\linewidth]{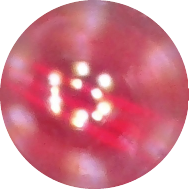}
        \caption{}
    \end{subfigure}
    \caption{\textbf{Properties of the mechanoresponsive photonic membrane.} (a) The membrane appears red when exposed to a red laser but prior to any contact. (b) Upon contact, the membrane undergoes a color change due to deformation. (c) For enhanced visualization, an augmented tactile imprint is created by applying the function \(f\) defined in \cref{eq:augmented_tactile_imprint}, using a specific setting of \(\alpha = 5\). (d) The membrane's speckled surface results from reflective glare when under illumination, necessitating the use of diffused lighting to reduce the effect.}
    \label{fig:contact}
\end{figure}

Taking into account the stringent requirements in terms of physical dimensions and sensitivity to deformation, we opt for a mechanoresponsive photonic membrane, as illustrated in Miller \etal~\cite{miller2022scalable}. Unlike conventional color-changeable membranes~\cite{yuan2017gelsight}, which necessitate complex and bulky lighting systems, this membrane boasts a remarkable feature: on deformation, it is susceptible to changes in reflection wavelength, leading to color change even under basic illumination. Images of the undeformed and deformed membrane are depicted in \cref{fig:contact}(a) and (b), respectively. We refer the readers to the original paper~\cite{miller2022scalable} for details of the principles that underlie the workings of this mechanoresponsive photonic membrane.

To enhance visualization of the color change, we introduce the concept of tactile imprint, representing the difference between images of deformed and undeformed membranes, denoted as \(\mathcal{I}_w\) and \(\mathcal{I}_n\) respectively. This imprint is further amplified by a constant factor \(\alpha\). The function \(f\) that yields the augmented tactile imprint is defined as follows:
\begin{equation}
    f(\mathcal{I}_n, \mathcal{I}_w) = g(\alpha(\mathcal{I}_w - \mathcal{I}_n)+\beta),
    \label{eq:augmented_tactile_imprint}
\end{equation}
where \(\beta\) is an offset, typically set to \(\frac{255}{2}\), to convert negative differences to positive values, and \(g\) is a clipping function ensuring the value falls within the image range \([0, 255]\), so as to be compatible with the RGB value range for visualization. The augmented tactile imprint of the contact shown in \cref{fig:contact}(b) can be seen, more clearly, in \cref{fig:contact}(c).

To fabricate the membrane, we use a commercially available holographic photopolymer, Litiholo C-RT20, and employ the Lippman photographic technique. Initially, we vertically expose the holographic photopolymer membrane, which has been peeled off from its initial support and layered onto a stainless-steel mirror sheet, in a darkroom under a \SI{5}{mW} red laser at a distance of \SI{40}{cm} for \SI{5}{min}. Following this exposure, the film exhibits a uniform red pattern generated from standing waves, resulting from the reflection of the red laser on the mirror sheet. These waves capture structural color patterns as periodic refractive index variations (distributed Bragg reflectors) within the material. The material then undergoes plasma surface treatment for \SI{1}{min} to enhance its surface adhesion. Recognizing the limited strength of the thin membrane, we utilize black silicone DOWSIL 700 as the soft backing layer to improve durability. The backing layer is laminated onto the treated side of the membrane, and the mixture is then cured in an oven at \(\SI{70}{\degreeCelsius}\) for 2 hours.

\paragraph*{Data acquisition system}

The ability to monitor subtle color changes in the mechanoresponsive photonic membrane at high spatial resolution is critical. For this purpose, we utilize an off-the-shelf miniaturized camera module equipped with an OV9734 CMOS image sensor. This sensor offers a resolution of \(1280 \times 720\) pixels, which is ideal for capturing the detailed images necessary for accurate analysis. Taking into account the ultra-compact size requirements of our sensor, the camera is paired with a lens that offers a \SI{76}{\degree} \ac{fov}, chosen to minimize optical distortion while fitting within the limited cross-sectional diameter of the sensor.

To maximize the fidelity of the color capture, the camera operates at full resolution using the uncompressed YUV format, which ensures that the images are detailed and free from compression artifacts. This setup not only enhances sensitivity to subtle color changes, but also contributes to the production of clearer, less noisy images. Additionally, the incorporation of USB connectivity in the data acquisition system allows for easy integration with standard computing devices, ensuring versatility and ease of use in a variety of applications.

\paragraph*{Lighting system and diffusing layer}

The mechanoresponsive photonic membrane, essential for our sensor's functionality, shifts its reflection wavelength upon contact. To capture this change accurately, we use six white surface-mounted LEDs arranged in a circular pattern around the camera lens, ensuring comprehensive coverage across all wavelengths.

Because the membrane's reflective surface can cause significant glare, (see illustration in \cref{fig:contact}(d)), we incorporate a diffusing layer made of translucent silicone, specifically Posilicone DRSGJ02, positioned directly behind the membrane. The chosen material not only surpasses traditional methods such as diffuser paper in terms of light distribution uniformity but also adheres naturally to the photonic membrane. This adhesive property is beneficial as it mitigates the use of additional bonding agents whose constituent chemicals could interact with the membrane causing color alterations or other unwanted issues. The translucent quality of the silicone ensures even light distribution, enhancing the sensor's accuracy in detecting and responding to color changes.

\paragraph*{Transparent silicone layer}

To ensure robust support and optimal functionality within the sensor's assembly, a transparent silicone layer, made from Smooth-on Solaris, is positioned between the camera and the diffusing layer. This material is prepared by mixing equal quantities of Part A and Part B and is then degassed using a vacuum pump and poured directly above the camera within the sensor's metal shell to form a stable base.

Once cured at room temperature after $24$ hours, a second layer of Posilicone DRSGJ02 silicone is mixed in the same 1:1 ratio, degassed using a vacuum pump, and poured over the base to create the diffusing layer, which helps to ensure uniform light distribution. The mechanoresponsive photonic membrane and its support structure are then placed on top of this layer, and the entire assembly is left to cure for $12$ hours at room temperature. This step-by-step process ensures that all components are chemically bonded, enhancing the sensor's structural integrity and functional performance with clear optical paths and effective light diffusion.

\paragraph*{Sensor shell and extended handle}

To enhance the sensor's robustness and durability, we designed a metal shell with an outer diameter of \SI{8}{mm} and a minimal wall thickness of \SI{0.5}{mm}, targeting a large sensing region during \ac{ramis}. This shell securely houses the camera module as shown in \cref{fig:sensor_design}. The lower end of the shell is attached to a carbon tube via a custom fixture, which acts as an extended handle. The design strengthens the sensor's structure while enhancing its usability and ergonomic handling, ultimately facilitating its seamless integration into \ac{ramis} systems.

\subsection{Calibration}\label{sec:calibration}

This section details the calibration process for \sensorname, which is crucial to ensure that it can extract accurate deformation data from sensor readings. We focus on converting pixel color changes into surface deformation data, utilizing the HSV color space as recommended by Miller \etal~\cite{miller2022scalable}. This color space has been shown to correlate strongly with the deformation of mechanoresponsive photonic elastomers.

\begin{figure}[t!]
    \centering
    \includegraphics[width=1\linewidth]{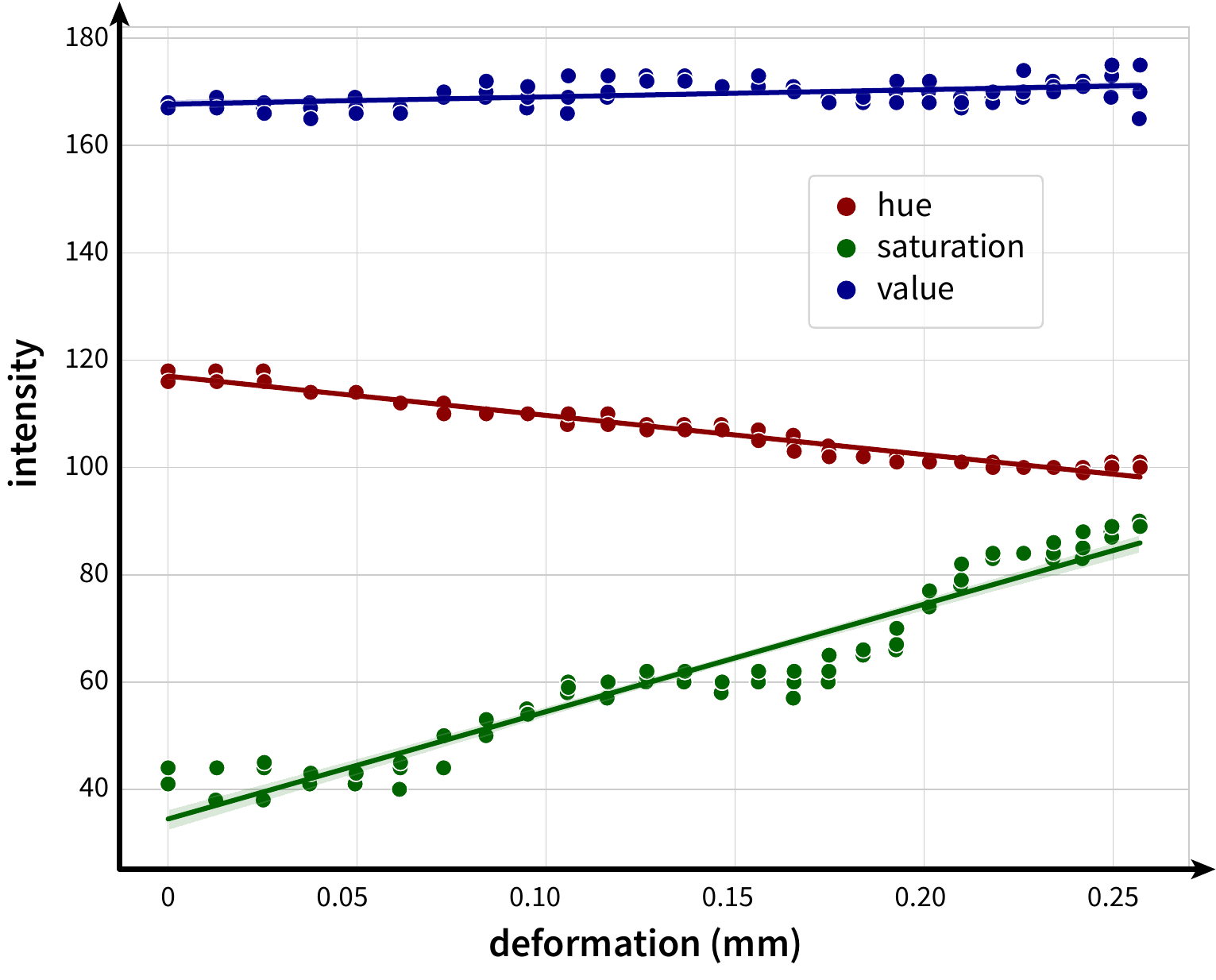}
    \caption{\textbf{Relationship between membrane deformation and its color.} The mechanoresponsive photonic membrane exhibits color changes that correlate directly with its deformation, providing a visual indication of mechanical stress.}
    \label{fig:hsv}
\end{figure}

We analyze and illustrate these correlations to understand the relationship between membrane deformation and the HSV channels, as shown in \cref{fig:hsv}. Ideally, a uniform membrane color would enable straightforward mapping from color variation \((\Delta H, \Delta S, \Delta V)\) to surface deformation \(D\) using a single function. However, real-world variations in membrane color, as observed in \cref{fig:contact}, necessitate a more complex approach. To address this, we introduce positional factors into our calibration model. We define a function \(m\) that maps color changes \((\Delta H(u, v), \Delta S(u, v), \Delta V(u, v))\) at each pixel position \((u,v)\) to the corresponding deformation depth. The function is mathematically formulated as follows:
\begin{equation}
    D(u, v) = m(\Delta H(u,v), \Delta S(u,v), \Delta V(u,v), u, v).
\end{equation}

For the practical implementation of this function, we use an \ac{mlp} to approximate the function \(m\). Following the recommendation by Wang \etal~\cite{wang2021gelsight}, we adopt a three-layer architecture (5-32-32-32-1), which has proven effective. This setup utilizes the hyperbolic tangent (\(\tanh\)) activation function, the Adam optimizer with a learning rate of \(0.001\), and \ac{mse} as the loss function. To train the network, we generate a dataset by applying a known-sized metal sphere to the sensor surface and recording the resulting color changes at the contact site. A dataset comprising 30 captures is sufficient to train a single sensor.

\begin{figure}[t!]
    \centering
    \includegraphics[width = 1\linewidth]{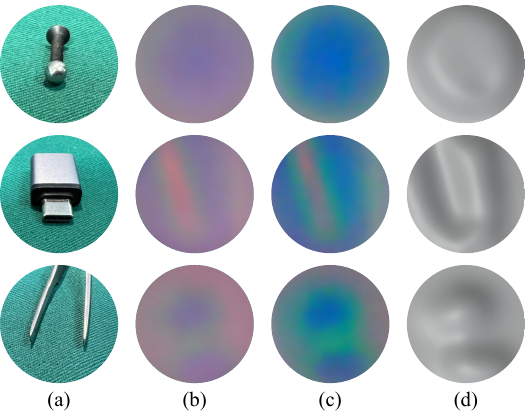}
    \caption{\textbf{Deformation reconstruction at the contact site obtained by \sensorname.} \sensorname analyzes color shifts within captured images to accurately reconstruct surface deformation at the contact site, offering high spatial resolution: (a) test objects (b) raw readings from \sensorname (c) augmented tactile imprints (d) final reconstructed deformations.}
    \label{fig:reconstruction}
\end{figure}

\cref{fig:reconstruction} presents qualitative results, illustrating the reconstructed sensor surface deformation from various kinds of contact, demonstrating our calibration method's efficacy.

\subsection{Sensor Sensitivity}\label{sec:sensor_sensitivity}

Once calibrated, \sensorname accurately measures surface deformations at the contact site. Its sensitivity to external force is critical to its ability to distinguish between soft and hard objects. Indeed the key capabilities required for robotic applications can be summarized as follows:
\begin{enumerate}
    \item The sensor must be able to detect deformations caused by minimal force in order to discern slight differences in hardness.
    \item It should identify incremental deformation differences caused by minor increases in force, enabling differentiation between similar hardness levels.
    \item The sensor should not reach saturation unless subjected to a relatively large force, so it retains the ability to discern substantial variations in hardness.
\end{enumerate}

To evaluate the sensitivity of \sensorname, we use an ATI Nano17 force sensor to apply controlled pressure to the sensor's surface and record its readings (\cref{fig:sensitivity}(a)). As shown in \cref{fig:sensitivity}(b), the sensor demonstrates high sensitivity, is capable of detecting deformations at \SI{0.02}{N}, can resolve force increments at approximately \SI{0.6}{mN}, and does not saturate under forces up to \SI{0.11}{N}.

\subsection{Sensor Repeatability and Hysteresis}\label{sec:sensor_hysteresis}

\begin{figure}[t!]
    \centering
    \begin{subfigure}[b]{.415\linewidth}
        \includegraphics[width=\linewidth]{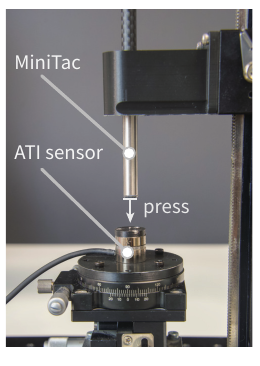}
        \caption{apparatus}
    \end{subfigure}%
    \hfill%
    \begin{subfigure}[b]{.58\linewidth}
        \includegraphics[width=\linewidth]{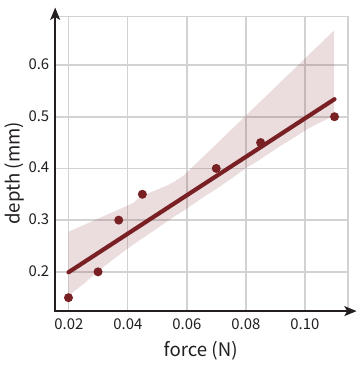}
        \caption{profile}
    \end{subfigure}
    \caption{\textbf{Sensitivity of \sensorname.} (a) The apparatus, equipped with an indenter, is used to measure deformation under varying external forces recorded by an ATI force sensor. (b) \sensorname detects deformation starting from \SI{0.02}{N}, resolves force increments at approximately \SI{0.6}{mN}, and remains unsaturated under force of up to \SI{0.11}{N}.}
    \label{fig:sensitivity}
\end{figure}

With the same setting in \cref{fig:sensitivity}(a), the repeatability of \sensorname \(r\) is calculated as:
\begin{equation}
r = \frac{{\Delta\hat{d}_t}_M}{d_M} \times 100\% = \frac{0.11}{0.5} = 22\%,
\end{equation}
where $d_M = 0.5~\mathrm{mm}$ is the maximum ground truth depth applied in this study and ${\Delta\hat{d}_t}_M = 0.11~\mathrm{mm}$ is the maximum measurement difference at the same step across all trials. The standard deviation of the difference image captured before loading and after unloading is $0.7$.

\begin{figure}[t!]
    \centering
    \includegraphics[width=\linewidth]{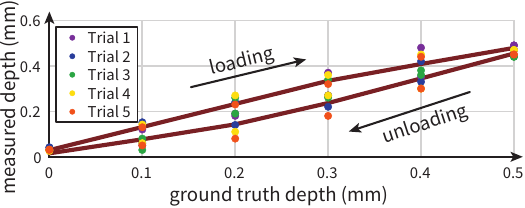}
    \caption{\textbf{Repeatability and hysteresis properties of \sensorname}. }
    \label{fig:repeatability}
\end{figure}

The red curve in \cref{fig:repeatability} is the smoothed average of the five trials measured above. Based on this curve, the maximum measurement difference between two processes $ {\Delta\hat{d}_p}_M$ is observed as $0.19~\mathrm{mm}$, and the hysteresis \(h\) is calculated as
\begin{equation}
h = \frac{{\Delta\hat{d}_p}_M}{d_M} \times 100\% = \frac{0.19}{0.5} = 38\%. 
\end{equation}

Given that the loading process yields output values higher than those of the unloading process, it is likely that this discrepancy is due to the viscoelastic properties of the elastomer.

\section{Experiments}\label{sec:exps}

Initially, we evaluate the efficacy of \sensorname in detecting hard embedded tumors using phantoms (\cref{sec:phantoms}), establishing a decision boundary so as to distinguish tumor tissue from normal tissue. Further validation is then performed ex-vivo to verify the efficacy of \sensorname as well as the established decision boundary (\cref{sec:ex-vivo}).

\begin{figure}[t!]
    \centering
    \includegraphics[width=1\linewidth]{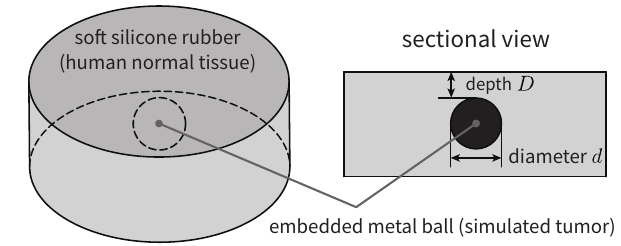}
    \caption{\textbf{Illustration of phantoms.} Phantoms made of soft silicone rubber mimic human tissue---in this case, a metal ball of diameter \(d\) is embedded at depth \(D\) to represent a hard tumor.}
    \label{fig:phantoms}
\end{figure}

\begin{figure}[t!]
    \centering
    \begin{subfigure}[b]{.5\linewidth}
        \includegraphics[width=\linewidth]{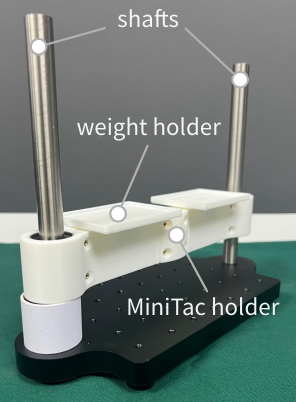}
        \caption{apparatus}
    \end{subfigure}%
    \begin{subfigure}[b]{.5\linewidth}
        \includegraphics[width=\linewidth]{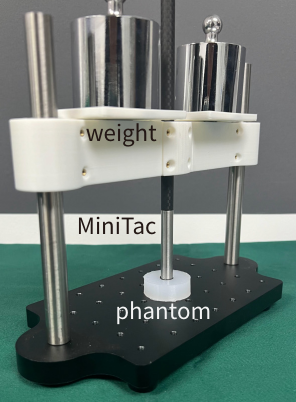}
        \caption{experiment with \sensorname}
    \end{subfigure}\\
    \begin{subfigure}[b]{\linewidth}
        \includegraphics[width=\linewidth]{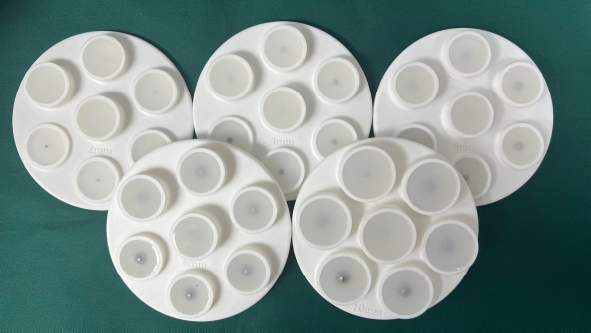}
        \caption{various phantoms}
    \end{subfigure}%
    \caption{\textbf{Experimental setup for \sensorname to detect embedded simulated tumors in phantoms.} (a) Side view of the device used to regulate the applied force. (b) \sensorname applying a consistent force of \SI{1000}{g} to the phantom. (c) Various phantoms are used to establish the discrimination boundary between those with and without embedded simulated tumors.}
    \label{fig:phantoms_platform}
\end{figure}

\begin{figure}[t!]
    \centering
    \begin{subfigure}[b]{\linewidth}
        \includegraphics[width=\linewidth]{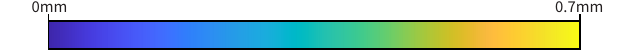}
        \caption{colorbar}
    \end{subfigure}\\%
    \begin{subfigure}[b]{\linewidth}
        \includegraphics[width=\linewidth]{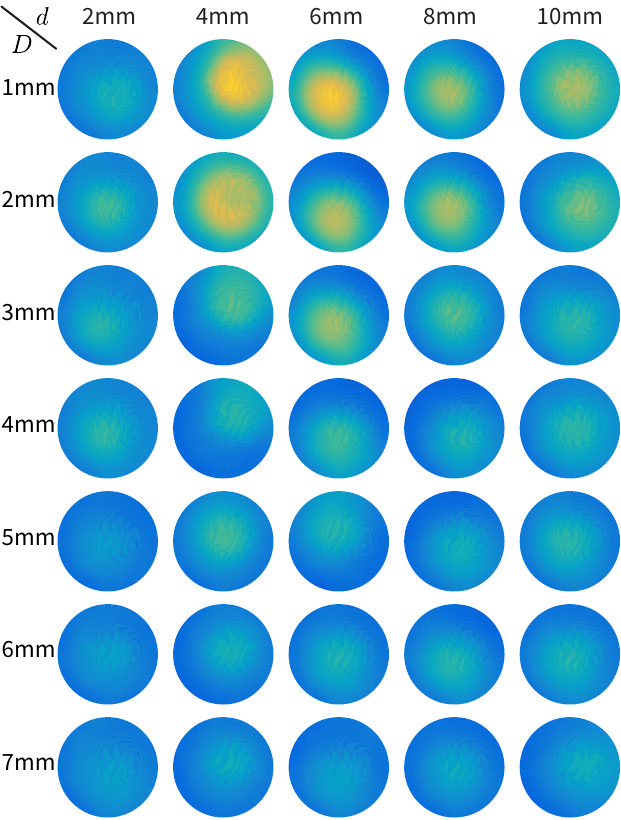}
        \caption{reconstructed deformations with embedded simulated tumors}
    \end{subfigure}\\%
    \begin{subfigure}[b]{\linewidth}
        \includegraphics[width=\linewidth]{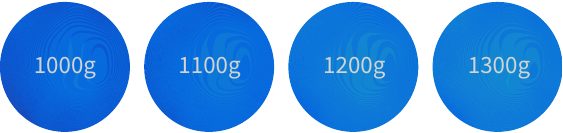}
        \caption{reconstructed deformations without embedded simulated tumors}
    \end{subfigure}
    \caption{\textbf{Reconstructed deformations of \sensorname when in contact with phantoms.} (a) Colorbar indicating the mapping from color to deformation value. (b) Reconstructed deformations obtained by applying a consistent force of \SI{1000}{g} to a selection of phantoms with embedded simulated tumors. (c) Reconstructed deformations obtained by applying varying forces, from \SI{1000}{g} to \SI{1300}{g}, to phantoms without embedded simulated tumors.}
    \label{fig:phantoms_results}
\end{figure}

\subsection{Experiments on Phantoms}\label{sec:phantoms}

We begin by evaluating the efficacy of \sensorname in detecting embedded hard tumors using controlled phantoms, aiming to establish a decision boundary between healthy and pathological (tumor-laden) tissue.

To create phantoms that closely resemble human tissue, we use Smooth-On Ecoflex 00-30 silicone rubber, which has a Shore 00 hardness of 30, similar to that of human tissue. The methodology follows the guidelines described in \cite{jia2013lump}. To assess the performance of \sensorname in a variety of tumor scenarios, we embed metal balls of assorted diameters \(d\) across different depths \(D\), simulating a range of embedded tumor situations. The phantom setup is illustrated in \cref{fig:phantoms}.

To establish a robust decision boundary between phantoms with and without simulated tumors, we collect samples under two distinct conditions:
\begin{itemize}[leftmargin=*,noitemsep,nolistsep]
    \item \textbf{Positive samples:} Phantoms are embedded with metal balls ranging in diameter from \SI{2}{mm} to \SI{10}{mm} and placed at depths from \SI{1}{mm} to \SI{7}{mm} at \SI{1}{mm} intervals. We apply a consistent force of \SI{1000}{g} on these samples, repeating the process four times for each. This relatively small force is chosen to generate fewer tactile features, thereby increasing the classification challenge to fashion a robust classifier. This force level is within the range that humans apply during manual palpation without causing discomfort.
    \item \textbf{Negative samples:} Phantoms without embedded metal balls are subjected to variable forces ranging from \SI{1000}{g} to \SI{1300}{g} at \SI{100}{g} intervals.  These increased forces, as compared to those used for positive samples, are intended to generate more tactile features for data augmentation, further complicating the classification task and further ensuring the robustness of the classifier. To balance the sample distribution, we apply each force to the empty phantom \(35\) times to equalize the number of positive and negative samples.
\end{itemize}

To precisely control the force applied during data collection, we construct a specialized device as depicted in \cref{fig:phantoms_platform}. It features a platform that securely holds the \sensorname in place and also allows for the placement of weights to regulate the applied force. The movement of the platform is constrained by two smooth vertical shafts, ensuring that the force is directed strictly downwards.

One example for each scenario is showcased in \cref{fig:phantoms_results}(b) for positive samples and \cref{fig:phantoms_results}(c) for negative samples, respectively. The data from positive samples (\cref{fig:phantoms_results}(b)) distinctly indicate the presence of embedded tumors, with larger diameters and shallower depths resulting in more pronounced deformation signatures. In contrast, \cref{fig:phantoms_results}(c) demonstrates that deformations in phantoms without tumors tend to have flatter features due to the uniformity of the phantom material.

Based on the collected data, we initially divided it into training and test sets in a \(4:1\) ratio. We then employ a \ac{svm} classifier with a linear kernel to distinguish between positive and negative samples, using the mean (\(\mu\)) and standard deviation (\(\sigma\)) of the deformation data as features. We apply standardization to normalize these features to zero mean and unit variance. The classifier achieves \(100\%\) accuracy in both training and test sets. The decision boundary defined by the \ac{svm} is given by:
\begin{equation}
    0.33z(\mu) + 4.80z(\sigma) = -4.53.
    \label{eq:db}
\end{equation}

\subsection{Experiments on Ex-Vivo Tissues}\label{sec:ex-vivo}

To further validate the performance of \sensorname and the robustness of the established decision boundary, we conduct tests on ex-vivo tissue. This setup introduces a more realistic setting that better simulates clinical conditions.

\begin{figure}[t!]
    \centering
    \begin{subfigure}[b]{\linewidth}
        \includegraphics[width=\linewidth]{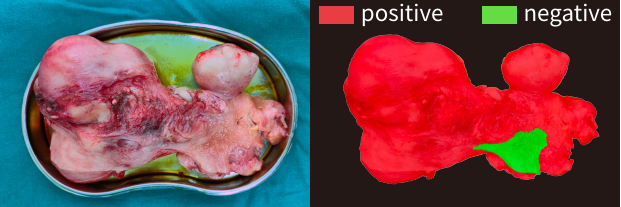}
        \caption{an ex-vivo sample with regions labeled by surgeons}
    \end{subfigure}\\%
    \begin{subfigure}[b]{\linewidth}
        \includegraphics[width=\linewidth]{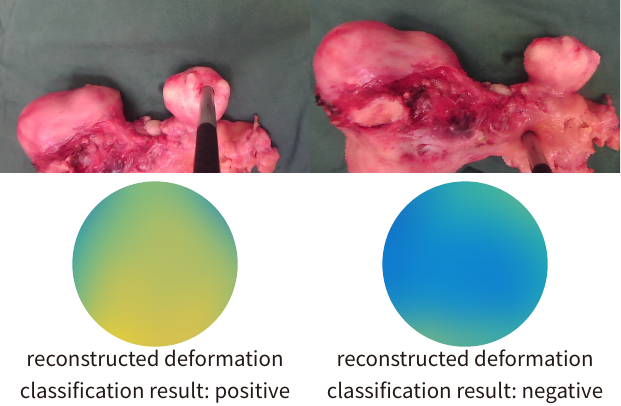}
        \caption{tumor detection by \sensorname}
    \end{subfigure}
    \caption{\textbf{Effectiveness of \sensorname in detecting embedded tumors in ex-vivo human tissue.} (a) An ex-vivo sample with regions labeled by surgeons, indicating the presence (positive) or absence (negative) of tumors. (b) The tumor detection results from \sensorname align with those of the surgeons, demonstrating the effectiveness of \sensorname.}
    \label{fig:ex-vivo}
\end{figure}

\cref{fig:ex-vivo} illustrates our ex-vivo testing process. In \cref{fig:ex-vivo}(a), we show an ex-vivo tissue sample with areas containing and lacking an embedded tumor, as identified by experienced surgeons through manual palpation. Using \sensorname, surgeons then performed 50 random presses across this tissue. Our established decision boundary (\cref{eq:db}) achieved 100\% accuracy in these trials, perfectly aligning with the surgeons' ground truth assessments. \cref{fig:ex-vivo}(b) provides two representative samples of reconstructed deformation and classification results. This strong correlation between \sensorname's outputs and expert surgical evaluations demonstrates the sensor's effectiveness in accurately detecting tumors within human tissue.

\section{Discussion}\label{sec:discussion}

This section highlights the unique benefits of \sensorname in enhancing palpation capabilities within \ac{ramis} systems. Unlike traditional tactile sensors such as piezoresistive~\cite{kalantari2010design}, piezoelectric~\cite{lee2014micro}, capacitive~\cite{kim2018sensorized}, and optical-fiber-based sensors~\cite{xie2014optical}, \sensorname integrates \num{300000} taxels within a \SI{8}{mm}-diameter area. This configuration allows for detailed surface mapping, effectively capturing widespread deformations.

The ability to accurately map deformations and calculate mean and variance has proven effective in the detection of hard objects such as tumors, as demonstrated in controlled phantom experiments (\cref{sec:phantoms}) and ex-vivo tissue applications (\cref{sec:ex-vivo}). Harder objects induce more pronounced deformations, enhancing detectability and increasing the deformation data's mean and variance. The decision boundary analysis confirms the importance of standard deviation, which remains significant even when larger forces are applied (\cref{eq:db}). Although similar results might be achieved with high-resolution vision-based tactile sensors such as GelSight\cite{jia2013lump} and DIGIT\cite{di2024using}, \sensorname's simple illumination system allows for a significantly more compact design, facilitating integration into \ac{ramis} systems (\cref{fig:teaser}).

\section{Conclusion}\label{sec:conclusion}

In this paper, we introduce \sensorname, an ultra-compact vision-based tactile sensor designed to enhance \ac{ramis} by providing high-resolution tissue palpation capabilities within the confined spaces of surgical environments. With its slender \SI{8}{mm} diameter, \sensorname integrates seamlessly into mainstream \ac{ramis} systems, such as the Da Vinci robotic surgical system, effectively overcoming the traditional limitations of tactile feedback. The sensor's ability to differentiate between normal and pathological tissue (tumors) has been validated through extensive testing on both phantoms and ex-vivo samples, demonstrating its sensitivity and efficacy.

While \sensorname's ultra-compact design shows promise for \ac{ramis} integration, further clinical trials are needed to assess its performance in live surgical settings and compatibility with established procedures. Nevertheless, \sensorname has the potential to improve surgical outcomes and advance medical robotics.

\paragraph*{Acknowledgment}

We extend our sincere thanks to Yuyang Li (Peking University) for his assistance with figure drawing, Lei Yan (LeapZenith AI Research) for mechanical design support, and Ms. Hailu Yang (Peking University) for her assistance in procuring every piece of raw material necessary for this research.

\balance
\bibliographystyle{IEEEtran}
\bibliography{reference_header,reference}
\end{document}